\documentclass[a4paper,twoside]{article}

\usepackage{epsfig}
\usepackage{subcaption}
\usepackage{calc}
\usepackage{amssymb}
\usepackage{amstext}
\usepackage{amsmath}
\usepackage{amsthm}
\usepackage{multicol}
\usepackage{pslatex}
\usepackage{apalike}
\usepackage{algorithm2e}
\usepackage{siunitx}
\usepackage{booktabs}
\usepackage{xcolor}
\usepackage[bottom]{footmisc}
\usepackage{orcidlink}

\usepackage{SCITEPRESS}     

\begin{document}
\title{Sampling in CMA-ES: Low Numbers of Low Discrepancy Points}

\author{
\authorname{
Jacob de Nobel\orcidlink{0000-0003-1169-1962}, 
Diederick Vermetten\orcidlink{0000-0003-3040-7162}, 
Thomas H.W. B{\"a}ck\orcidlink{0000-0001-6768-1478}, 
Anna V. Kononova\orcidlink{0000-0002-4138-7024}
}
\affiliation{LIACS, Leiden University, Leiden, Netherlands}
\email{\{j.p.de.nobel, d.l.vermetten, t.h.w.baeck, a.kononova\}@liacs.leidenuniv.nl}
}


\abstract{
The Covariance Matrix Adaptation Evolution Strategy (CMA-ES) is one of the most successful examples of a derandomized evolution strategy. However, it still relies on randomly sampling offspring, which can be done via a uniform distribution and subsequently transforming into the required Gaussian. Previous work has shown that replacing this uniform sampling with a low-discrepancy sampler, such as Halton or Sobol sequences, can improve performance over a wide set of problems. We show that iterating through small, fixed sets of low-discrepancy points can still perform better than the default uniform distribution. Moreover, using only 128 points throughout the search is sufficient to closely approximate the empirical performance of using the complete pseudorandom sequence up to dimensionality 40 on the BBOB benchmark. For lower dimensionalities (below 10), we find that using as little as 32 unique low discrepancy points performs similar or better than uniform sampling. In 2D, for which we have highly optimized low discrepancy samples available, we demonstrate that using these points yields the highest empirical performance and requires only 16 samples to improve over uniform sampling. Overall, we establish a clear relation between the $L_2$ discrepancy of the used point set and the empirical performance of the CMA-ES.
}

\onecolumn \maketitle \normalsize \setcounter{footnote}{0} \vfill

\section{Introduction}
Optimization techniques play a crucial role in various scientific and engineering applications. Exact methods systematically explore the parameter space but often suffer from inefficiency due to their exhaustive nature. For example, it has been shown that randomized search is superior to grid search for hyperparameter tuning~\cite{bergstra2012random}. This is because, especially in higher dimensions, a randomized process will provide better coverage of sample points in the domain than an exhaustive search, given a limited evaluation budget. While samples generated uniformly at random provide an improvement in exploring the search space, such samples can still be quite suboptimal in covering the domain. Given a limited number of samples, uniform samples can be distributed very unevenly~\cite{halton1960efficiency}. This notion of evenly spreading points across a given domain motivates the research into \emph{Low-Discrepancy Sequences}. These are sequences of pseudo-randomly generated points that are designed to minimize the gaps and clusters that often occur in uniform random sampling, providing more uniform coverage of the search space. Specifically, discrepancy measures are designed to measure how regularly a given point set is distributed in a given space~\cite{clement2023computing}. While early work with low discrepancy point sets focuses on Monte Carlo integration~\cite{halton1960efficiency,sobol1967distribution}, they have subsequently been used in various domains, such as computer vision~\cite{paulin2022matbuilder} and financial modeling \cite{galanti1997low}. Low Discrepancy point sets have been used in the optimization domain to set up the Design of Experiments (DoE) within a constrained budget~\cite{santner2003doe}. One application of particular interest in our context is one-shot optimization, where low-discrepancy sequences have been shown to outperform more traditional uniform sampling~\cite{bousquet2017critical}.
Moreover, random search using quasi-random points has been shown to outperform traditional random search~\cite{niederreiter1992random}. In metaheuristics, randomized search is employed by many different algorithms, such as Evolution Strategies (ES) \cite{Beyer01}. In ES, quasi-random point sets have been used as an alternative sampling strategy to pure random sampling~\cite{teytaud2007dcma}, specifically for the CMA-ES~\cite{hansen2001completely}. Simplified, the procedure of the CMA-ES can be divided into two steps, which are repeated until convergence: 

\begin{itemize}
    \item Sample $\lambda$ points from a multivariate normal distribution $\mathcal{N}(\mathbf{m}, \mathbf{C})$.
    \item Adjust the parameters of the multivariate normal distribution to move towards the $\mu$ points with the highest fitness.    
\end{itemize}
Modifying the sampling step to use points from a pseudo-random sequence~\cite{teytaud2015quasi} demonstrated increased performance and stability on benchmark functions. Moreover, this furthers  \emph{derandomization}, which aims to achieve self-adaptation without any independent stochastic variation of the strategy parameters~\cite{back2023thirty}. This paper aims to extend this work by focusing on the number of samples drawn from a (pseudo) random sampling strategy specifically for the CMA-ES. Specifically, we investigate if repeatedly reusing small subsets of pseudorandom sequences in a deterministic manner can be an effective sampling strategy for the CMA-ES. We provide an analysis for several well-known low discrepancy point sets to investigate the relation between discrepancy and empirical performance on the BBOB benchmark functions. 

\section{Preliminaries}
\label{sec:preliminaries}

\subsection{Low-discrepancy sequences}
The discrepancy of a set of points quantifies how regularly they are spaced in the domain. One of the most common discrepancy measures of a point set $P\subseteq[0,1]^d$ is the $L_{\infty}$ star discrepancy~\cite{clement2023constructing}, which is defined as follows:
\begin{equation}
   d^*_{\infty}(P) = \textit{sup}_{q\in[0,1]^d}\left|\frac{|P\cap[0,q)|}{|P|}-\lambda(q)\right|    
\end{equation}
Here, $\lambda(q)$ is the Lebesgu measure of the box $[0,q)$ and $d^*_{\infty}(P)$ measures the worst absolute difference between $\lambda(q)$ of a d-dimensional box anchored at the origin and the proportion of points that fall inside this box. Note that this measure should be minimized to evenly space points in the domain. Since the $L_\infty$ star discrepancy can be computationally expensive~\cite{clement2023computing}, we can also consider the $L_2$ star discrepancy, which is defined as follows~\cite{zhou2013mixture}:
\begin{equation} 
    d^*_{2}(P) =\left( \int_{[0,1]^d}\left|\frac{|P\cap[0,q)|}{|P|}-\lambda(q)\right| \,d  q\right)^{1/2} 
\end{equation}
Several pseudo-random sequences have lower star discrepancies than corresponding uniform sequences. These methods include Latin Hypercube Sampling~\cite{lhs}, Jittered sampling~\cite{jittered}, and Hammersly sequences~\cite{hammersly}. Our work considers the Halton~\cite{halton1960efficiency} and Sobol~\cite{sobol1967distribution} sequences. 

\subsection{Derandomization and CMA-ES}
While Evolution Strategies~\cite{Beyer01} depend on a random process to sample candidate solutions, the internal parameter update has been \emph{derandomized} in state-of-the-art implementations of the algorithm. Derandomization ensures self-adaptation happens without any independent stochastic variation of the strategy parameters~\cite{ostermeier1994derandomized}. Effectively, this means the update of the strategy parameters is decoupled from the sampling of candidate solutions, moving away from the notion that `good solutions have good parameters' of traditional ES. This allows modern ES, such as the CMA-ES, to be more robust and learn good strategy parameters while using relatively small population sizes~\cite{hansen2001completely}. Within the CMA-ES, the sampling procedure is the only remaining source of stochasticity. At every generation, $\lambda$ individuals are sampled from the $d$-dimensional Gaussian distribution $\mathcal{N}(\mathbf{m}, \sigma^2 \mathbf{C})$. Where $\mathbf{m}$ is the mean of the sampling distribution, $\sigma$ is the global step size, and $\mathbf{C}$ is the covariance matrix. In practice, $\mathbf{C}$ is spectrally decomposed into two matrices, $\mathbf{B}$ and $\mathbf{D}$, representing the eigenvectors and inverse square root of the eigenvalues of $\mathbf{C}$, respectively. This allows the sampling of points in the CMA-ES to happen in a three-stage process:
\begin{enumerate}
    \item $\mathbf{z_k} \sim \mathcal{N}(\mathbf{0}, \mathbf{1})$
    \item $\mathbf{y_k} = \mathbf{B} \mathbf{D} \mathbf{z_k} \quad \quad  \sim \mathcal{N}(\mathbf{0}, \mathbf{C})$
    \item $\mathbf{x_k} = \mathbf{m} + \sigma \mathbf{y_k}  \quad  \sim \mathcal{N}(\mathbf{m}, \sigma^2\mathbf{C})$
\end{enumerate}
Given this decomposition, the first step of the sampling process can be practically achieved by sampling from a uniform distribution $\mathbf{u} \sim \mathcal{U}(0,1)^d$, and transforming each coordinate $\mathbf{u}_i$ of the sample by: 
\begin{equation}
\phi^{-1}(\mathbf{u}_i) = \sqrt{2} \text{erf}^{-1}(2\mathbf{u}_i - 1)
\end{equation}
which is the inverse of the cumulative density function for a standard Gaussian distribution. 
  
Since the sampling procedure in CMA-ES can be seen as sampling in $[0,1]^d$, replacing the uniform sampling with a low-discrepancy sequence is a natural step. Previous work has shown that scrambled Halton sequences can improve performance over the standard sampling procedure on most considered benchmark problems~\cite{teytaud2007dcma,teytaud2015quasi}.

\section{Methods}
\subsection{Point set generation}
For our experiments, we modify how the CMA-ES samples from a normal distribution by exchanging the uniform number generator with a selection of points from a fixed point set stored in a cache. The cache is randomly permuted once, at the beginning of an optimization run, and cycled through in steps of size $\lambda$. The permutation is performed to break the bias that might be present in the ordering of the point set.

\begin{figure}[t!]
    \centering
    \includegraphics[width=0.45\textwidth]{figures/log10\_l2\_discrepancy.pdf}
    \caption{(Average) $\log_{10}(d^*_2)$ star discrepancy for the generated fixed-size point sets across all dimensionalities. Colors are (min-max) normalized on a per-dimensionality basis; darker colors indicate a worse (higher) $d_2^*$ value.}
    \label{fig:discrepancy}
\end{figure} 

We consider four methods for generating point sets: Sobol, Halton, uniform, and `optimized'. Halton sequences are known to have unwanted correlations in higher dimensional spaces, and as such, we employ a scrambling method to prevent this~\cite{braaten1979}. For the Sobol sequences, its balance properties require that the number of points generated is equal to a power of 2, so we always round up our number of points sampled to the closest power of 2. Finally, our `optimized' method for generating low-discrepancy point sets is split into two parts based on search space dimensionality. When $d=2$, we use optimized Fibonacci sets~\cite{clement2023constructing}, considered among the best low-discrepancy point sets available. However, these are only available for 2D since generating optimized low-discrepancy point sets is a very hard computational problem.
We use the improved Threshold Accepting subset selection heuristic from~\cite{clement2024heuristic} for higher dimensionalities, using Sobol sequences as the base sets. 

We create point sets of sizes $k \in \{16,32,64,128,256\}$ for each generation mechanism and calculate their $L_2$ star discrepancy\footnote{We use $L_2$ instead of $L_\infty$ star discrepancy for computational reasons.}\footnote{For the uniform points sets, we calculate the average over 100 samples.}. Figure~\ref{fig:discrepancy} shows how the discrepancy changes with increasing dimensionality and size of the point sets. We observe that, as expected, the standard uniform sampling has a noticeably higher discrepancy than both Halton and Sobol sequences, with the optimized method usually having the lowest discrepancy. Note that for the optimized method, the relative difference in discrepancy is largest when $d=2$ due to the different generation mechanism used for this dimensionality.  

\begin{figure}[t]
    \centering
    \includegraphics[width=0.45\textwidth]{figures/eaf\_auc\_per\_dim.pdf}
    \caption{Average area under the EAF curve for each sampling method on the BBOB benchmark, grouped by dimension. Colors are (min-max) normalized on a per-dimensionality basis; darker colors indicate a worse (lower) EAF value.}
    \label{fig:eaf_auc}
\end{figure}

\begin{figure*}[t!]
    \centering
    \includegraphics[width=0.95\textwidth]{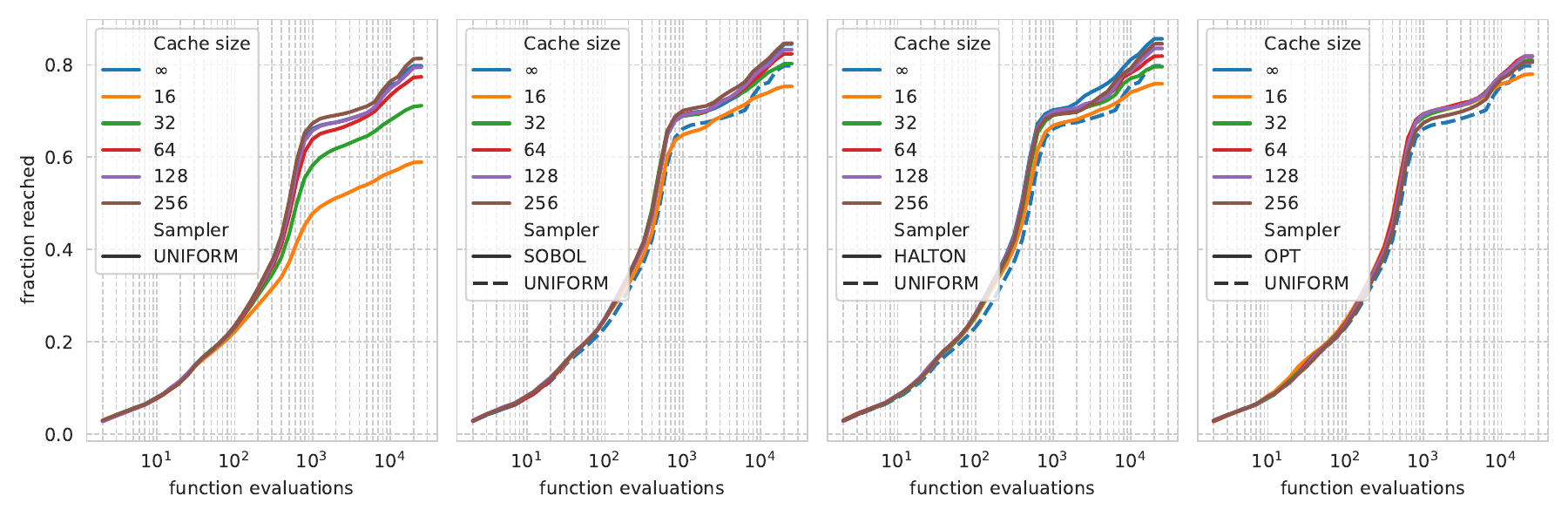}
    \caption{Empirical Attainment Function aggregated over all 24 BBOB functions for dimensionality 2. The methods are shown in color for each sampling strategy, with cache size indicating the number of points in the cache and $\infty$ indicating no caching; every sample is unique. From left to right, the subfigures show results using a uniform sampler, a Sobol sequence, a scrambled Halton sequence, and the optimized point sets. Note that in the three rightmost figures, the default CMA-ES sampling strategy, UNIFORM-$\infty$, is included for comparison. }
    \label{fig:eafd2}
\end{figure*}

\subsection{Experimental setup}
To gauge the impact of our derandomization, we run the Modular CMA-ES~\cite{modcma} with each of the generated point sets on the single-objective, noiseless BBOB suite~\cite{bbobfunctions}. This suite consists of 24 functions, of which we use $100$ different instances (generated by transformations such as rotation and translation) in dimensionalities $d\in\{2,5,10,20,40\}$. We set our evaluation budget at $B=d\cdot 10\,000$. By using only a single run on each instance, we only need to generate a single point set for each $(d, N)$-pair, while the fact that we use $100$ instances ensures a large enough sample size for each function. Additionally, we benchmark the default sampling mechanism, UNIFORM-$\infty$, and sampling from arbitrarily long scrambled Halton and Sobol sequences, denoted by the $\infty$ suffix in our results. No caching is applied in these cases, and each generated sample in the optimization process is unique. 

To measure performance, we consider the attainment-based cumulative distribution function~\cite{eaf}, with bounds $10^2$ and $10^{-8}$ for the log-scaled precision values.

\subsubsection*{Reproducibility}
All used point sets, the full experimental code, and data are in our reproducibility repository~\cite{reproducibility_and_figures}.

\section{Results}

Figure~\ref{fig:eaf_auc} shows each sampling method's normalized area under the EAF curve. Without caching, we observe that both low-discrepancy sequences outperform the uniform sampler, and overall, the scrambled Halton shows the highest empirical performance. Note that the differences tend to decrease with dimensionality. When comparing the results within each fixed point set size, the uniform sampler often performs notably worse than any of the low-discrepancy sets. This effect is especially noticeable for smaller set sizes (lower values of $k$) but remains observable for all cache sizes. Interestingly, the optimized point sets only seem to bring performance benefits in low dimensionalities and using smaller cache sizes. For higher dimensionalities, they often perform similarly to uniform sampling. Overall, increasing the size of the cached point set increases performance, and using the complete sequence ($\infty$) yields the highest performance for each method. Interestingly, point sets of only 64 to 128 samples are often enough to get very close in performance to using the complete sequence as a sampling strategy. Moreover, we find that such sequences still outperform the default sampling strategy of the CMA-ES, i.e., UNIFORM-$\infty$, especially in lower dimensions. In $2D$, we can observe that only using 16 optimized samples can outperform the default sampling strategy.

\begin{figure}[!h]
    \centering
    \includegraphics[width=0.5\textwidth]{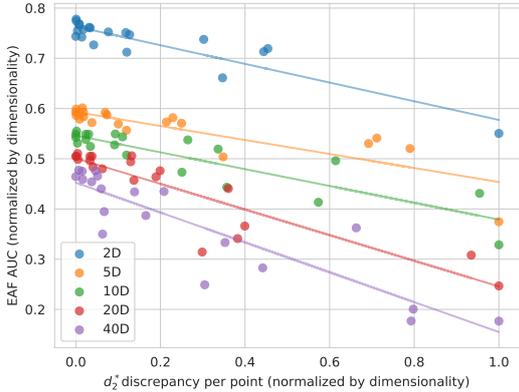}
    \caption{Average area under the Empirical Attainment Function over all BBOB functions, grouped by dimension vs. the $L_2$ star discrepancy, normalized by dimensionality and the number of points. Lines indicate a linear (least-squares) model for each dimensionality.}
    \label{fig:eaf_corr}
\end{figure}

When looking at the performance of the algorithm variants over time, as visualized in Figure~\ref{fig:eafd2}, we notice that the differences between the different sampling methods occur rather early in the search. While the used set does not have a noticeable impact on the initialization, soon after, the small sets with low discrepancy are seen to stagnate, resulting in the much worse anytime performance observed in Figure~\ref{fig:eaf_auc}. By comparing the different sampler to the dashed line of the default uniform sequence, we note that all sets of size 32 and higher outperform this baseline at the end of the optimization. Additionally, we should note that the optimized sets perform best among the smallest sample sizes, but the Halton and Sobol sequences benefit more from increasing their size. Plots for other dimensionalities are available in the appendix. 
Finally, we show the relation between discrepancy and performance in Figure~\ref{fig:eaf_corr}, where we observe a clear correlation between these two measures. This supports our previous observation that low discrepancy point sets are beneficial to the performance of the CMA-ES. 

\subsection{Multiples of $\lambda$}
\begin{figure}[h]
    \centering
    \includegraphics[width=0.47\textwidth]{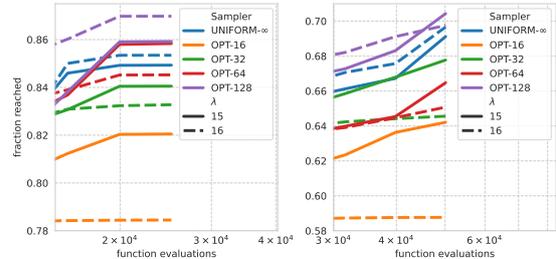}
    \caption{Empirical Attainment Function aggregated over all 24 BBOB functions for dimensionality 2 (left) and 5 (right), zoomed to final fraction reached. The default sampling strategy of the CMA-ES, UNIFORM-$\infty$, is shown in comparison to using a cached sampling strategy, which uses the `OPT' samples for a cache size $k \in \{16, 32, 64, 128\}$. The solid lines represent $\lambda = 15$ and the dashed lines $\lambda = 16$.}
    \label{fig:eaf_lambda}
\end{figure}

Since the CMA-ES only requires $\lambda$ points at every iteration, a natural assumption would be that only $\lambda$ unique points would be required for an effective sampling strategy, given that these points cover the domain sufficiently. In our previous experiments, we have used the default setting for $\lambda$, i.e., $4 + \lfloor 3 + \text{ln}(d) \rfloor$, which is strictly less than 16, the smallest tested cache size $k$. This causes the sampler to cycle through the point set, causing variance between the samples used in each generation. Here, we investigate this effect using either a population size $\lambda$ of 15 or 16. In the latter case, $\lambda$ is a multiple of the cache size, and for $k = 16$, this ensures that every generation uses exactly the same samples. For $k < 16$, this causes a cyclic pattern, where every $\frac{k}{16}$-th generation has the same samples. 
In figure~\ref{fig:eaf_lambda}, the empirical performance of this experiment is visualized. The figure shows that when using the default sampling strategy, i.e., UNIFORM-$\infty$, using a $\lambda = 16$ is better than using $\lambda = 15$. This is reversed for a cache size $k = 16$, where there is no variance between the samples in subsequent generations. For larger $k$, this performance of $\lambda = 15$ becomes closer to $\lambda = 16$. This indicates there must be variation between generations, not only between the samples within each generation, which aligns with~\cite{teytaud2007dcma}. Intuitively, this makes sense, as the strategy parameters are being averaged over multiple generations, making diversity between subsequent populations important. Notably, though, using only 4-8 unique populations during the entire optimization procedure already yields higher performance than the UNIFORM-$\infty$ sampling strategy (for $d=2$ and $d=5$).  

\section{Conclusions and Future Work}
In this paper, we have shown that replacing the uniform sampling with low discrepancy samples in the offspring generation of the CMA-ES can yield clear benefits in performance on commonly used benchmarks. While this matches previous observations on the advantage of low-discrepancy sequences~\cite{teytaud2015quasi}, our experiments on repeatedly (re-)using small sets of points show that we don't necessarily need to rely on generators when sampling with the CMA-ES. While we note that, in general, sampling from an arbitrarily long low discrepancy sequence is better than using a fixed point set, such sets remain competitive when using only 128 points. In fact, on the two-dimensional problems, where our $L_\infty$-optimized point sets can be considered state-of-the-art~\cite{clement2023constructing}, a cached set consisting of only 16 samples is enough to outperform the default sampling strategy of the CMA-ES. 

However, our results in higher dimensionalities show that while there is a clear correlation between the discrepancy of a point set and the performance of the CMA-ES using it, this correlation is not perfect. While our optimized sets generated by improved Threshold Accptance~\cite{clement2024heuristic} have lower discrepancy than the corresponding Halton and Sobol sets, their anytime performance is slightly worse, indicating that there might be other aspects of the point sets we should take into account. 

We have shown that not only does the discrepancy of samples within a single generation impact performance, but the diversity between subsequent generations also has an impact. Future work might focus on diving deeper into the relationship between point set size, discrepancy, diversity, and performance.

\paragraph*{Acknowledgements}
We want to thank Fran{\c{c}}ois Cl{\'e}ment and Carola Doerr for providing us with the optimized low-discrepancy point sets used in this paper.

\bibliographystyle{apalike}
{\small
\bibliography{refs}}

\begin{thebibliography}{}

\bibitem[Bergstra and Bengio, 2012]{bergstra2012random}
Bergstra, J. and Bengio, Y. (2012).
\newblock Random search for hyper-parameter optimization.
\newblock {\em Journal of machine learning research}, 13(2).

\bibitem[Beyer, 2001]{Beyer01}
Beyer, H. (2001).
\newblock {\em {The theory of evolution strategies}}.
\newblock Natural computing series. Springer.

\bibitem[Bousquet et~al., 2017]{bousquet2017critical}
Bousquet, O., Gelly, S., Kurach, K., Teytaud, O., and Vincent, D. (2017).
\newblock Critical hyper-parameters: No random, no cry.
\newblock {\em arXiv preprint arXiv:1706.03200}.

\bibitem[Braaten and Weller, 1979]{braaten1979}
Braaten, E. and Weller, G. (1979).
\newblock An improved low-discrepancy sequence for multidimensional quasi-monte carlo integration.
\newblock {\em Journal of Computational Physics}, 33(2):249--258.

\bibitem[Bäck et~al., 2023]{back2023thirty}
Bäck, T. H.~W., Kononova, A.~V., van Stein, B., Wang, H., Antonov, K.~A., Kalkreuth, R.~T., de~Nobel, J., Vermetten, D., de~Winter, R., and Ye, F. (2023).
\newblock {Evolutionary Algorithms for Parameter Optimization—Thirty Years Later}.
\newblock {\em Evolutionary Computation}, 31(2):81--122.

\bibitem[Cl{\'e}ment et~al., 2023a]{clement2023constructing}
Cl{\'e}ment, F., Doerr, C., Klamroth, K., and Paquete, L. (2023a).
\newblock Constructing optimal $l_{\infty}$ star discrepancy sets.
\newblock {\em arXiv preprint arXiv:2311.17463}.

\bibitem[Cl{\'e}ment et~al., 2024]{clement2024heuristic}
Cl{\'e}ment, F., Doerr, C., and Paquete, L. (2024).
\newblock Heuristic approaches to obtain low-discrepancy point sets via subset selection.
\newblock {\em Journal of Complexity}, 83:101852.

\bibitem[Cl{\'e}ment et~al., 2023b]{clement2023computing}
Cl{\'e}ment, F., Vermetten, D., De~Nobel, J., Jesus, A.~D., Paquete, L., and Doerr, C. (2023b).
\newblock Computing star discrepancies with numerical black-box optimization algorithms.
\newblock In {\em Proceedings of the Genetic and Evolutionary Computation Conference}, pages 1330--1338.

\bibitem[de~Nobel et~al., 2024]{reproducibility_and_figures}
de~Nobel, J., Vermetten, D., Kononova, A.~V., and B{\"a}ck, T. (2024).
\newblock Reproducibility files and additional figures.

\bibitem[de~Nobel et~al., 2021]{modcma}
de~Nobel, J., Vermetten, D., Wang, H., Doerr, C., and B{\"{a}}ck, T. (2021).
\newblock Tuning as a means of assessing the benefits of new ideas in interplay with existing algorithmic modules.
\newblock In Krawiec, K., editor, {\em {GECCO} '21: Genetic and Evolutionary Computation Conference, Companion Volume, Lille, France, July 10-14, 2021}, pages 1375--1384. {ACM}.

\bibitem[Galanti and Jung, 1997]{galanti1997low}
Galanti, S. and Jung, A. (1997).
\newblock Low-discrepancy sequences: Monte carlo simulation of option prices.
\newblock {\em The Journal of Derivatives}, 5(1):63--83.

\bibitem[Halton, 1960]{halton1960efficiency}
Halton, J.~H. (1960).
\newblock On the efficiency of certain quasi-random sequences of points in evaluating multi-dimensional integrals.
\newblock {\em Numerische Mathematik}, 2:84--90.

\bibitem[Hansen et~al., 2009]{bbobfunctions}
Hansen, N., Finck, S., Ros, R., and Auger, A. (2009).
\newblock Real-parameter black-box optimization benchmarking 2009: Noiseless functions definitions.
\newblock Research Report RR-6829, {INRIA}.

\bibitem[Hansen and Ostermeier, 2001]{hansen2001completely}
Hansen, N. and Ostermeier, A. (2001).
\newblock Completely derandomized self-adaptation in evolution strategies.
\newblock {\em Evolutionary computation}, 9(2):159--195.

\bibitem[Loh, 1996]{lhs}
Loh, W.-L. (1996).
\newblock On latin hypercube sampling.
\newblock {\em The annals of statistics}, 24(5):2058--2080.

\bibitem[L{\'{o}}pez{-}Ib{\'{a}}{\~{n}}ez et~al., 2024]{eaf}
L{\'{o}}pez{-}Ib{\'{a}}{\~{n}}ez, M., Vermetten, D., Dr{\'{e}}o, J., and Doerr, C. (2024).
\newblock Using the empirical attainment function for analyzing single-objective black-box optimization algorithms.
\newblock {\em CoRR}, abs/2404.02031.

\bibitem[Niederreiter, 1992]{niederreiter1992random}
Niederreiter, H. (1992).
\newblock {\em Random number generation and quasi-Monte Carlo methods}.
\newblock SIAM.

\bibitem[Ostermeier et~al., 1994]{ostermeier1994derandomized}
Ostermeier, A., Gawelczyk, A., and Hansen, N. (1994).
\newblock A derandomized approach to self-adaptation of evolution strategies.
\newblock {\em Evolutionary Computation}, 2(4):369--380.

\bibitem[Paulin et~al., 2022]{paulin2022matbuilder}
Paulin, L., Bonneel, N., Coeurjoly, D., Iehl, J.-C., Keller, A., and Ostromoukhov, V. (2022).
\newblock {MatBuilder}: Mastering sampling uniformity over projections.
\newblock {\em ACM Transactions on Graphics (proceedings of SIGGRAPH)}.

\bibitem[Pausinger and Steinerberger, 2016]{jittered}
Pausinger, F. and Steinerberger, S. (2016).
\newblock On the discrepancy of jittered sampling.
\newblock {\em Journal of Complexity}, 33:199--216.

\bibitem[Peart, 1982]{hammersly}
Peart, P. (1982).
\newblock The dispersion of the hammersley sequence in the unit square.
\newblock {\em Monatshefte f{\"u}r Mathematik}, 94(3):249--261.

\bibitem[Santner et~al., 2003]{santner2003doe}
Santner, T., Williams, B., and Notz, W. (2003).
\newblock {\em The Design and Analysis of Computer Experiments}.
\newblock Springer Series in Statistics, Springer.

\bibitem[Sobol', 1967]{sobol1967distribution}
Sobol', I.~M. (1967).
\newblock On the distribution of points in a cube and the approximate evaluation of integrals.
\newblock {\em Zhurnal Vychislitel'noi Matematiki i Matematicheskoi Fiziki}, 7(4):784--802.

\bibitem[Teytaud, 2015]{teytaud2015quasi}
Teytaud, O. (2015).
\newblock Quasi-random numbers improve the {CMA-ES} on the {BBOB} testbed.
\newblock In Bonnevay, S., Legrand, P., Monmarch{\'{e}}, N., Lutton, E., and Schoenauer, M., editors, {\em Artificial Evolution - 12th International Conference, Evolution Artificielle, {EA} 2015, Lyon, France, October 26-28, 2015. Revised Selected Papers}, volume 9554 of {\em Lecture Notes in Computer Science}, pages 58--70. Springer.

\bibitem[Teytaud and Gelly, 2007]{teytaud2007dcma}
Teytaud, O. and Gelly, S. (2007).
\newblock {DCMA:} yet another derandomization in covariance-matrix-adaptation.
\newblock In Lipson, H., editor, {\em Genetic and Evolutionary Computation Conference, {GECCO} 2007, Proceedings, London, England, UK, July 7-11, 2007}, pages 955--963. {ACM}.

\bibitem[Zhou et~al., 2013]{zhou2013mixture}
Zhou, Y.-D., Fang, K.-T., and Ning, J.-H. (2013).
\newblock Mixture discrepancy for quasi-random point sets.
\newblock {\em Journal of Complexity}, 29(3-4):283--301.

\end{thebibliography}

\begin{appendix}

\begin{figure*}[t!]
\centering
\begin{subfigure}[t]{\textwidth}
    \centering
    \caption{Dimensionality 5}
    \includegraphics[width=0.95\textwidth]{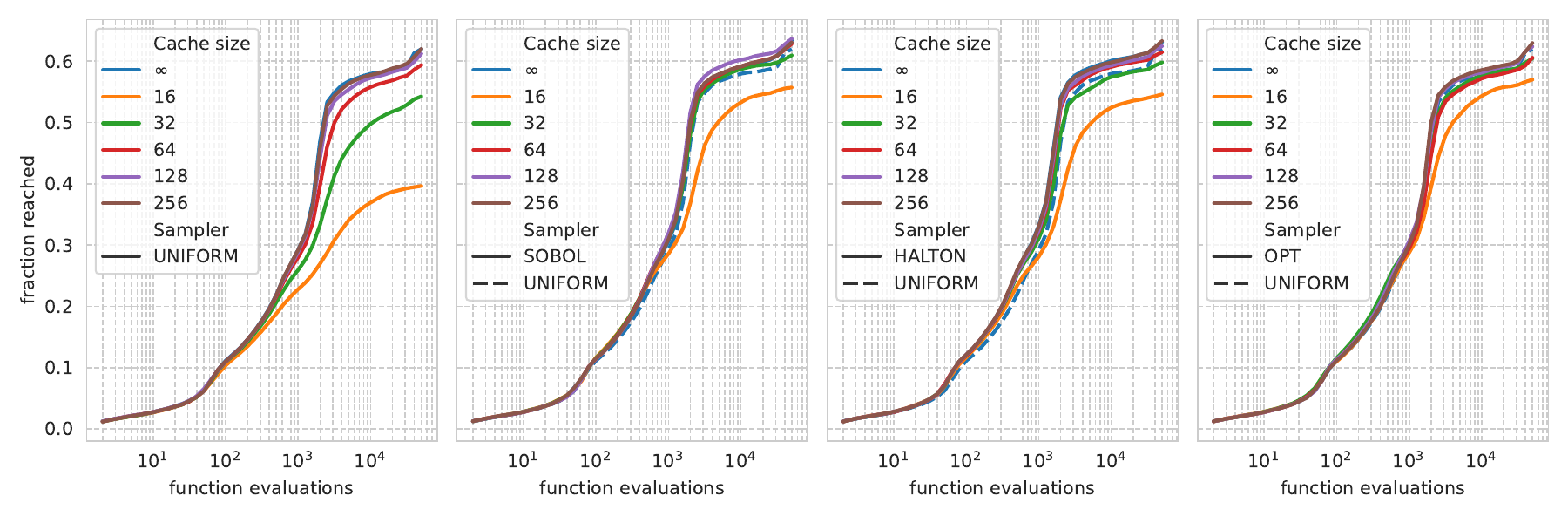}
    \label{fig:eafd5}
\end{subfigure}
~
\begin{subfigure}[t]{\textwidth}
    \centering
    \caption{Dimensionality 10}
    \includegraphics[width=0.95\textwidth]{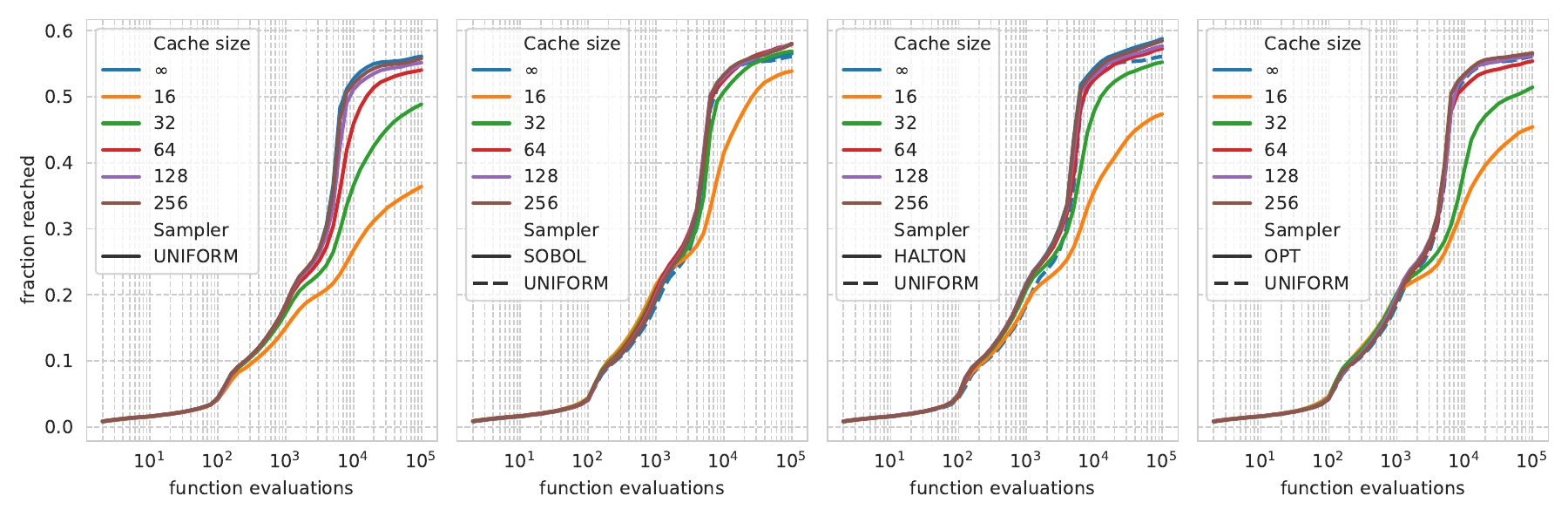}
    \label{fig:eafd10}
\end{subfigure}
~
\begin{subfigure}[t]{\textwidth}
    \centering
    \caption{Dimensionality 20}
    \includegraphics[width=0.95\textwidth]{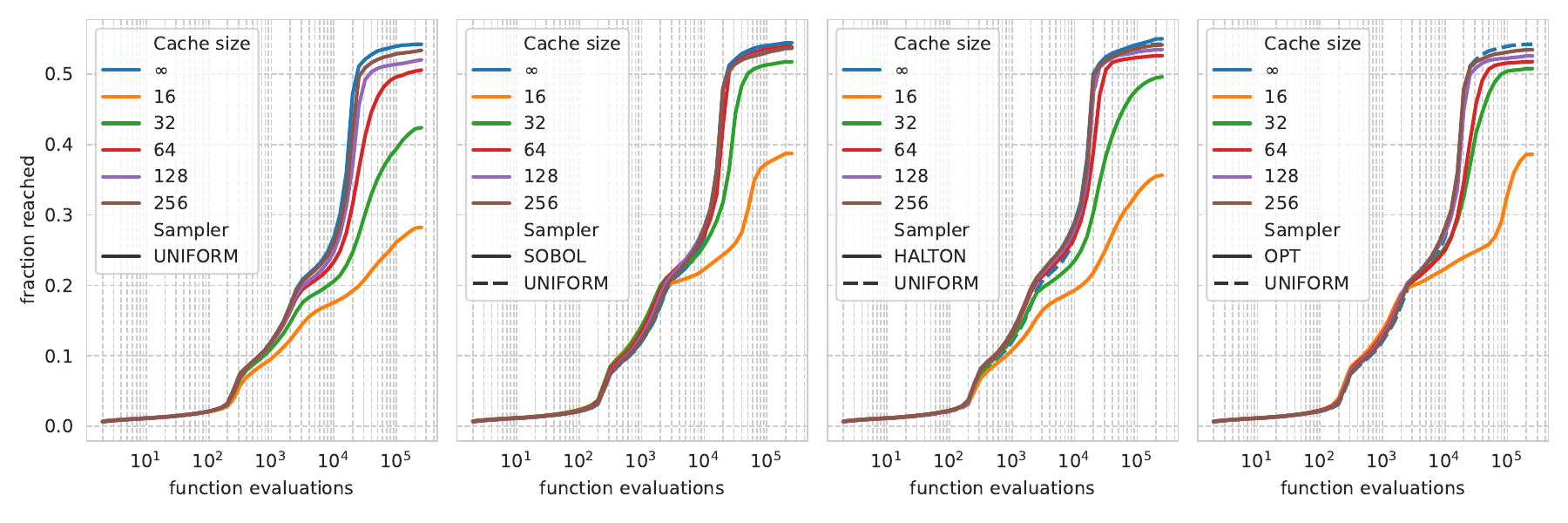}
    \label{fig:eafd20}
\end{subfigure}
~
\begin{subfigure}[t]{\textwidth}
    \centering
    \caption{Dimensionality 40}
    \includegraphics[width=0.95\textwidth]{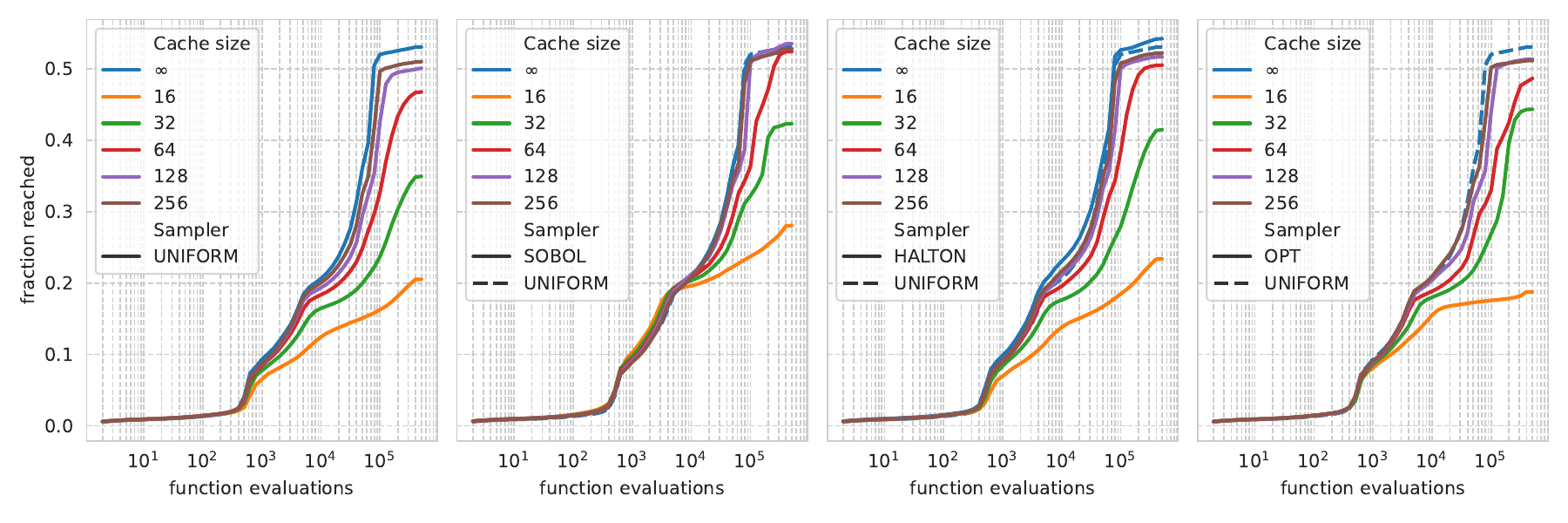}
    \label{fig:eafd40}
\end{subfigure}
\caption{Empirical Attainment Function aggregated over all 24 BBOB functions for dimensionalities 5, 10, 20, and 40 (from top to bottom). 
The methods are shown in color for each sampling strategy, with cache size indicating the number of points in the cache and $\infty$ indicating no caching; every sample is unique. From left to right, the subfigures show results using a uniform sampler, a Sobol sequence, a scrambled Halton sequence, and the optimized point sets. Note that in the three rightmost figures, the default CMA-ES sampling strategy, UNIFORM-$\infty$, is included for comparison. 
}

\end{figure*}

\end{appendix}
\end{document}